\documentclass{article} 
\usepackage{comment}
\usepackage{nips13submit_e,times}
\usepackage{hyperref}
\usepackage{url}
\usepackage{natbib}
\usepackage{graphicx}
\usepackage{amsmath}
\usepackage{caption}
\usepackage[compact]{titlesec}

\title{Multi-digit Number Recognition from Street View Imagery using Deep Convolutional Neural Networks}

\author{
Ian J. Goodfellow, Yaroslav Bulatov, Julian Ibarz, Sacha Arnoud, Vinay Shet\\
Street View and reCAPTCHA Teams, Google Inc.\\
\texttt{[goodfellow,yaroslavvb,julianibarz,sacha,vinayshet]@google.com} \\
}

\nipsfinalcopy 

\begin{document}

\maketitle

\begin{abstract}
Recognizing arbitrary multi-character text in unconstrained natural photographs is a hard problem. In this paper, we address an equally hard sub-problem in this domain viz. recognizing arbitrary multi-digit numbers from Street View imagery. Traditional approaches to solve this problem typically separate out the localization, segmentation, and recognition steps. In this paper we propose a unified approach that integrates these three steps via the use of a deep convolutional neural network that operates directly on the image pixels. We employ the DistBelief~\citep{Dean-et-al-NIPS2012} implementation of deep neural networks in order to train large, distributed neural networks on high quality images. We find that the performance of this approach increases with the depth of the convolutional network, with the best performance occurring in the deepest architecture we trained, with eleven hidden layers. We evaluate this approach on the publicly available SVHN dataset and achieve over 96\% accuracy in recognizing complete street numbers. We show that on a per-digit recognition task, we improve upon the state-of-the-art, achieving 97.84\% accuracy. We also evaluate this approach on an even more challenging dataset generated from Street View imagery containing several tens of millions of street number annotations and achieve over 90\% accuracy. To further explore the applicability of the proposed system to broader text recognition tasks, we apply it to transcribing synthetic distorted text from a popular CAPTCHA service, reCAPTCHA. reCAPTCHA is one of the most secure reverse turing tests that uses distorted text as one of the cues to distinguish humans from bots. With the proposed approach we report a 99.8\% accuracy on transcribing the hardest category of reCAPTCHA puzzles. Our evaluations on both tasks, the street number recognition as well as reCAPTCHA puzzle transcription, indicate that at specific operating thresholds, the performance of the proposed system is comparable to, and in some cases exceeds, that of human operators. 
\end{abstract}

\section{Introduction}

Recognizing multi-digit numbers in photographs captured at street level is an important component of modern-day map making. A classic example of a corpus of such street level photographs is Google's Street View imagery comprised of hundreds of millions of geo-located 360 degree panoramic images.
The ability to automatically transcribe an address number from a geo-located patch of pixels and associate the
transcribed number with a known street address helps pinpoint, with a high degree of accuracy, the location of the building it represents.  

More broadly, recognizing numbers in photographs is a problem of interest to the optical character recognition community. While OCR on constrained domains like document processing is well studied, arbitrary multi-character text recognition in photographs is still highly challenging. This difficulty arises due to the wide variability in the visual appearance of text in the wild on account of a large range of fonts, colors, styles, orientations, and character arrangements. The recognition problem is further complicated by environmental factors such as lighting, shadows, specularities, and occlusions as well as by image acquisition factors such as resolution, motion, and focus blurs. 

In this paper, we focus on recognizing multi-digit numbers from Street View panoramas.
While this reduces the space of characters that need to be recognized, the complexities listed above still apply to this sub-domain. Due to these complexities, traditional approaches to solve this problem typically separate out the localization, segmentation, and recognition steps.
   
In this paper we propose a unified approach that integrates these three steps via the use of a deep convolutional
neural network that operates directly on the image pixels.
This model is configured with multiple hidden layers (our best configuration had eleven layers, but our experiments
suggest deeper architectures may obtain better accuracy, with diminishing returns), all with feedforward
connections.
We employ DistBelief to implement these large-scale deep neural networks. 

We have evaluated this approach on the publicly available Street View House Numbers (SVHN) dataset and achieve over 96\% accuracy in recognizing street numbers. We show that on a per-digit recognition task, we improve upon the state-of-the-art and achieve 97.84\% accuracy. We also evaluated this approach on an even more challenging dataset generated from Street View imagery containing several tens of millions of street number annotations and achieve over 90\% accuracy. Our evaluations further indicate that at specific operating thresholds, the performance of the proposed system is comparable to that of human operators. To date, our system has helped us extract close to 100 million street numbers from Street View imagery worldwide.

While the challenges listed above for numbers in Street View data can be considered to be real-world, natural variabilities in text, another class of data where text is deliberatly distorted synthetically is in CAPTCHA puzzles. CAPTCHAs~\cite{} are reverse turing tests designed to use distorted text to distinguish humans and machines running automated text recognition software. Synthetic distortions on these text based puzzles is used to increase variability in the visual appearance of the text, thus increasing transcription difficulty. In order to evaluate the general applicability of the peoposed approach to the broader task of recognizing arbitrary text, we applied it to the task of solving CAPTCHA puzzles from reCAPTCHA, one of the widely used CAPTCHA service on the internet. We show that we are able to achieve a 99.8\% accuracy on the hardest reCAPTCHA puzzle. 

The key contributions of this paper are:
(a) a unified model to localize, segment, and recognize multi-digit numbers from street level photographs
(b) a new kind of output layer, providing a conditional probabilistic model of sequences
(c) empirical results that show this model performing best with a deep architecture
(d) results of applying proposed model on the harderst category of reCAPTCHA images to achieve 99.8\% transcription accuracy
(e) reaching human level performance at specific operating thresholds.

%

\section{Related work}
\label{sec:related}

Convolutional neural networks~\citep{Fukushima80,LeCun98-small} are neural networks with sets of neurons
having tied parameters.
Like most neural networks, they contain several filtering layers with each layer applying
an affine transformation to the vector input followed by an elementwise non-linearity.
In the case of convolutional networks, the affine transformation can be implemented as a
discrete convolution rather than a fully general matrix multiplication. This makes convolutional
networks computationally efficient, allowing them to scale to large images.
It also builds equivariance to translation into the model
(in other words, if the image is shifted by one pixel to the right, then the output of the
convolution is also shifted one pixel to the right; the two representations vary equally with translation).
Image-based convolutional networks typically use a {\em pooling layer} which summarizes 
the activations of many adjacent filters with a single response. Such pooling layers may summarize
the activations of groups of units with a function such as their maximum, mean, or L2 norm.
These pooling layers help the network be robust to small translations of the input.

Increases in the availability of computational resources, increases in the size of available training sets, and
algorithmic advances such as the use of piecewise linear units~\citep{Jarrett-ICCV2009,Glorot+al-AI-2011,Goodfellow-et-al-ICML2013}
and dropout training~\citep{Hinton-et-al-arxiv2012} have resulted in many recent successes using deep convolutional
neural networks. \citet{Krizhevsky-2012-small} obtained dramatic improvements in the state of the art in object recognition.
\citet{Zeiler-Fergus-arxiv2013} later improved upon these results.

On huge datasets, such as those used at Google, overfitting is not an issue, and increasing the size of the network
increases both training and testing accuracy. To this end, \citet{Dean-et-al-NIPS2012} developed DistBelief,
a scalable implementation of deep neural networks, which includes support for convolutional networks.
We use this infrastructure as the basis for the experiments in this paper.

Convolutional neural networks have previously been used mostly for applications such as recognition of single
objects in the input image. In some cases they have been used as components of systems that solve more complicated
tasks. \citet{Girshick-et-al-arxiv2013} use convolutional neural networks as feature extractors for a system that
performs object detection and localization. However, the system as a whole is larger than the neural network portion
trained with backprop, and has special code for handling much of the mechanics such as proposing candidate object
regions. \citet{Szegedy-nips2013} showed that a neural network could learn to output a heatmap that could be
post-processed to solve the object localization problem. In our work, we take a similar approach, but with less
post-processing and with the additional requirement that the output be an ordered sequence rather than an unordered
list of detected objects.
\citet{Alsharif-Pineau-arxiv2013} use convolutional maxout networks~\citep{Goodfellow-et-al-ICML2013} to
provide many of the conditional probability distributions used in a larger model using HMMs to transcribe text from
images. In this work, we propose to solve similar tasks involving localization and segmentation, but we propose to
perform the entire task completely within the learned convolutional network. In our approach, there is no need for
a separate component of the system to propose candidate segmentations or provide a higher level model of the image.

\section{Problem description}
\label{sec:problem}

\begin{figure}
\centering
\begin{tabular}{cc}
\includegraphics[height=1.0in]{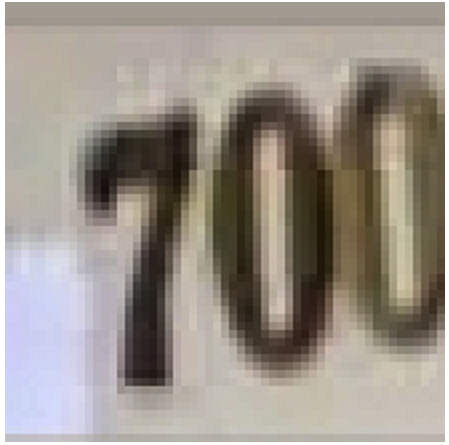} &%
\includegraphics[height=1.5in]{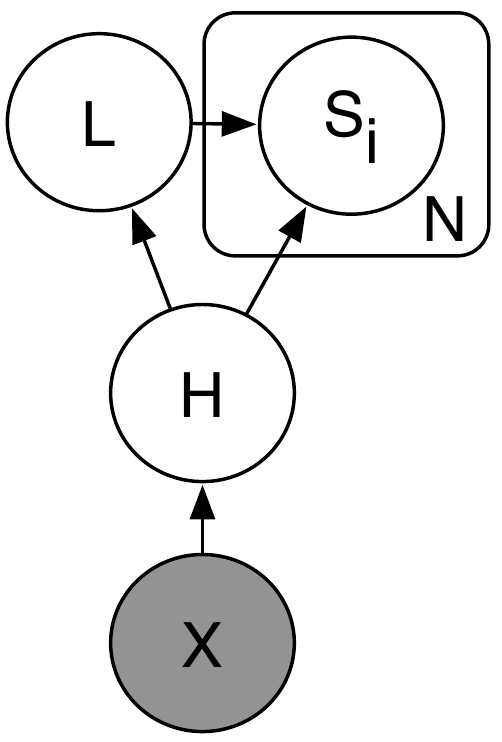} \\
a) & b)\\
\end{tabular}
\caption{a) An example input image to be transcribed. The correct output for this image is ``700''. 
b) The graphical model structure of our sequence transcription model, depicted
using plate notation~\citep{Buntine94} to represent the multiple $S_i$. Note that the relationship
between $X$ and $H$ is deterministic. The edges going from $L$ to $S_i$ are optional,
but help draw attention to the fact that our definition of $P(\mathbf{S} \mid X)$ does not
query $S_i$ for $i > L$.}
\label{hacky_fig}
\end{figure}

Street number transcription is a special kind of sequence recognition. Given an image, the task
is to identify the number in the image. See an example in Fig.~\ref{hacky_fig}a. The
number to be identified is a sequence of digits, $\mathbf{s} = s_1, s_2, \dots, s_n$. When determining
the {\em accuracy} of a digit transcriber, we compute the proportion of the input images for
which the length $n$ of the sequence and every element $s_i$ of the sequence is predicted correctly.
There is no ``partial credit'' for getting individual digits of the sequence correct. This is
because for the purpose of making a map, a building can only be found on the map from its address
if the whole street number was transcribed correctly.

For the purpose of building a map, it is extremely important to have at least human level accuracy.
Users of maps find it very time consuming and frustrating to be led to the wrong location, so it is
essential to minimize the amount of incorrect transcriptions entered into the map. It is, however,
acceptable not to transcribe every input image. Because each street number may have been photographed many
times, it is still quite likely that the proportion of buildings we can place on the map is greater than
the proportion of images we can transcribe. We therefore advocate evaluating this task based on the
{\em coverage} at certain levels of accuracy, rather than evaluating only the total degree of accuracy
of the system. To evaluate coverage, the system must return a confidence value, such as the probability
of the most likely prediction being correct. Transcriptions below some confidence threshold can then
be discarded. The coverage is defined to be the proportion of inputs that are not discarded. The
coverage at a certain specific accuracy level is the coverage that results when the confidence threshold
is chosen to achieve that desired accuracy level. For map-making purposes, we are primarily interested
in coverage at 98\% accuracy or better, since this roughly corresponds to human accuracy.

Using confidence thresholding allows us to improve maps incrementally over time--if we develop a system
with poor accuracy overall but good accuracy at some threshold, we can make a map with partial coverage,
then improve the coverage when we get a more accurate transcription system in the future. We can also
use confidence thresholding to do as much of the work as possible via the automated system and do the
rest using more expensive means such as hiring human operators to transcribe the remaining difficult inputs.

One special property of the street number transcription problem is that the sequences are of bounded
length. Very few street numbers contain more than five digits, so we can use models that assume the
sequence length $n$ is at most some constant $N$, with $N=5$ for this work. Systems that make such
an assumption should be able to identify whenever this assumption is violated and refuse to return a
transcription so that the few street numbers of length greater than $N$ are not incorrectly added to
the map after being transcribed as being length $N$. (Alternately, one can return the most likely
sequence of length $N$, and because the probability of that transcription being correct is low, the
default confidence thresholding mechanism will usually reject such transcriptions without needing
special code for handling the excess length case)

\section{Methods}
\label{sec:methods}

Our basic approach is to train a probabilistic model of sequences given images. Let $\mathbf{S}$ represent
the output sequence and $X$ represent the input image. Our goal is then to learn a model of $P(\mathbf{S} \mid X)$
by maximizing $\log P(\mathbf{S} \mid X)$ on the training set.

To model $\mathbf{S}$, we define $\mathbf{S}$ as a collection of $N$ random variables $S_1, \dots, S_N$ representing
the elements of the sequence and an additional random variable $L$ representing the length of the sequence.
We assume that the identities of the separate digits are independent from each other, so that the probability of a specific
sequence $\mathbf{s} = s_1, \dots, s_n$ is given by
\[ P(\mathbf{S} = \mathbf{s} | X) = P(L = n \mid X) \Pi_{i=1}^n P(S_i = s_i \mid X). \]
This model can be extended to
detect when our assumption that
the sequence has length at most $N$ is violated. To allow for detecting this case, we simply add an additional
value of $L$ that represents this outcome.

Each of the variables above is discrete, and when applied to the street number transcription problem, each has a small number
of possible values: $L$ has only 7 values (0, \dots, 5, and ``more than 5''), and each of the digit variables has 10 possible values.
This means it is feasible to represent each of them with a softmax classifier that receives as input
features extracted from $X$ by a convolutional neural network. We can represent these features as a random variable $H$
whose value is deterministic given $X$. In this model, $P(\mathbf{S} \mid X) = P(\mathbf{S} \mid H).$ See Fig.~\ref{hacky_fig}b for a graphical
model depiction of the network structure.

To train the model, one can maximize $\log P(\mathbf{S} \mid X)$ on the training set using a generic method like stochastic
gradient descent. Each of the softmax models (the model for $L$ and each $S_i$)
can use exactly the same backprop learning rule as when training an isolated softmax layer,
except that a digit classifier softmax model backprops nothing on examples for which that
digit is not present.

At test time, we predict
\[\mathbf{s} = (l, s_1, \dots, s_l) =  \text{argmax}_{L, S_1, \dots, S_L} \log P(S \mid X).\]
This argmax can be computed in
linear time. The argmax for each character can be computed independently. We then incrementally add up
the log probabilities for each character. For each length $l$, the complete log probability is given
by this running sum of character log probabilities, plus $\log P(l \mid x)$. The total runtime is thus $O(N).$

We preprocess by subtracting
the mean of each image. We do not use any whitening~\citep{Hyvarinen-2001-small},
local contrast normalization~\citep{sermanet-icpr-12}, etc.

\section{Experiments}
\label{sec:experiments}

In this section we present our experimental results. First, we describe our state of the
art results on the public Street View House Numbers dataset in section~\ref{subsec:public}.
Next, we describe the performance of this system on our more challenging, larger but internal
version of the dataset in section~\ref{subsec:internal}. We then present some experiments
analyzing the performance of the system in section~\ref{subsec:analysis}.

\subsection{Public Street View House Numbers dataset}
\label{subsec:public}

The Street View House Numbers (SVHN) dataset~\citep{Netzer-wkshp-2011} is a dataset of about 200k street numbers, along with bounding boxes for individual digits, giving about 600k digits total. To our knowledge, all previously published work cropped individual digits and tried to recognize those. We instead take original images containing multiple digits, and focus on recognizing them all simultaneously.

We preprocess the dataset in the following way -- first we find the small rectangular bounding box that will contain individual character bounding boxes.
We then expand this bounding box by 30\% in both the $x$ and the $y$ direction, crop the image to that bounding box and resize the crop to $64 \times 64$ pixels. We then crop a $54 \times 54$ pixel image from a
random location within the $64 \times 64$ pixel image. This means we generated several randomly shifted versions
of each training example, in order to increase the size of the dataset. Without this data augmentation, we lose
about half a percentage point of accuracy.
Because of the differing number of characters
in the image, this introduces considerable scale variability -- for a single digit street number, the digit fills the whole box, meanwhile a 5
digit street number will have to be shrunk considerably in order to fit.

Our best model obtained a sequence transcription accuracy of 96.03\%. This is not accurate enough to
use for adding street numbers to geographic location databases for placement on maps.
However, using confidence thresholding we obtain 95.64\% coverage at 98\% accuracy.
Since 98\% accuracy is the performance of human operators, these transcriptions are acceptable to include
in a map. We encourage researchers who work on this dataset in the future to publish coverage at 98\% accuracy
as well as the standard accuracy measure. Our system achieves a character-level accuracy of 97.84\%. This is
slightly better than the previous state of the art for a single network on the individual character task of
97.53\% \citep{Goodfellow-et-al-ICML2013}.

Training this model took approximately six days using 10 replicas in DistBelief. The exact training time varies
for each of the performance measures reported above--we picked the best stopping point for each performance measure
separately, using a validation set.

Our best architecture consists of eight convolutional hidden layers, one locally connected hidden layer, and two densely 
connected hidden layers. All connections are feedforward and go from one layer to the next (no skip connections). 
The first hidden layer contains maxout units~\citep{Goodfellow-et-al-ICML2013} (with three filters per unit) while
the others contain rectifier units~\citep{Jarrett-ICCV2009,Glorot+al-AI-2011}.
The number of units at each spatial location in each layer is
[48, 64, 128, 160] for the first four layers and 192 for all other locally connected layers. The fully connected layers
contain 3,072 units each.
Each convolutional layer includes max pooling and subtractive normalization. The max pooling window size is $2 \times 2$.
The stride alternates between 2 and 1 at each layer, so that half of the layers don't reduce the spatial size of
the representation. All convolutions use zero padding on the input to preserve representation size. The subtractive
normalization operates on 3x3 windows and preserves representation size. All convolution kernels were of size $5 \times 5$.
We trained with dropout applied to all hidden layers but not the input. 

\subsection{Internal Street View data}
\label{subsec:internal}

Internally, we have a dataset with tens of millions of transcribed street numbers.
However, on this dataset, there are no ground truth bounding boxes available.
We use an automated method (beyond the scope of this paper) to estimate the centroid of
each house number, then crop to a 128 $\times$ 128 pixel region surrounding the house
number. We do not rescale the image because we do not know the extent of the house number.
This means the network must be robust to a wider variation of scales than our public SVHN
network. On this dataset, the network must also localize the house number, rather than
merely localizing the digits within each house number.
Also, because the training
set is larger in this setting, we did not need augment the data with random translations.

\begin{figure}
\vspace{-2.5mm}
\begin{centering}
\includegraphics[width=.75\textwidth]{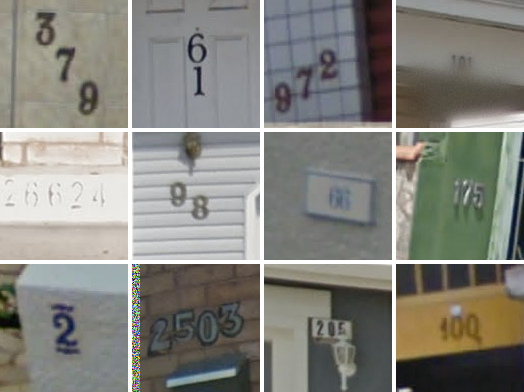}
\caption{Difficult but correctly transcribed examples from the internal street numbers dataset.
Some of the challenges in this dataset include diagonal or vertical layouts, incorrectly
applied blurring from license plate detection pipelines, shadows and other occlusions.
}
\label{fig:success}
\end{centering}
\end{figure}

\begin{figure}
\begin{centering}
\begin{tabular}{cccc}
\includegraphics[width=.2\textwidth]{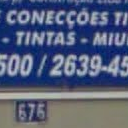} & \includegraphics[width=.2\textwidth]{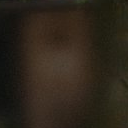}
& \includegraphics[width=.2\textwidth]{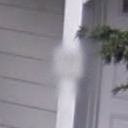} & \includegraphics[width=.2\textwidth]{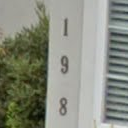} \\
100 vs. 676& 1110 vs. 2641 & 23 vs. 37 & 1 vs. 198 \\
\includegraphics[width=.2\textwidth]{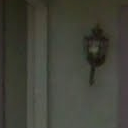} & \includegraphics[width=.2\textwidth]{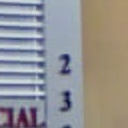}
& \includegraphics[width=.2\textwidth]{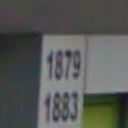} & \includegraphics[width=.2\textwidth]{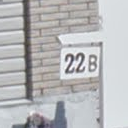} \\
4 vs. 332 & 2 vs 239 & 1879 vs. 1879-1883 & 228 vs. 22B \\
\includegraphics[width=.2\textwidth]{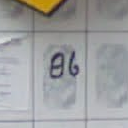} & \includegraphics[width=.2\textwidth]{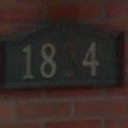}
& \includegraphics[width=.2\textwidth]{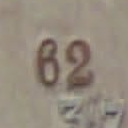} & \includegraphics[width=.2\textwidth]{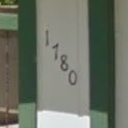} \\
96 vs. 86 & 1844 vs. 184 & 62 vs. 62-37 & 1180 vs. 1780 \\
\end{tabular}
\caption{
Examples of incorrectly transcribed street numbers from the large internal dataset (transcription vs. ground truth). Note that for
some of these, the ``ground truth'' is also incorrect.
The ground truth labels in this dataset are quite noisy, as is common in real world settings.
Some reasons for the ground truth errors in this dataset include:
1. The data was repurposed from an existing indexing pipeline where operators manually entered street
numbers they saw. It was impractical to use the same size of images as the humans saw, so heuristics
were used to create smaller crops. Sometimes the resulting crop omits some digits.
2. Some examples are fundamentally ambiguous, for instance street numbers including non-digit characters, or having multiple street numbers in same image which humans transcribed as a single number with an arbitrary separator like ``,'' or ``-''.
}
\label{fig:failure}
\end{centering}
\end{figure}

This dataset is more difficult because it comes from more countries (more than 12),
has street numbers with non-digit characters and the quality of the ground truth is lower.
See Fig.~\ref{fig:success} for some examples of difficult inputs from this dataset that our
system was able to transcribe correctly, and Fig.~\ref{fig:failure} for some examples of
difficult inputs that were considered errors.

We obtained an overall sequence transcription accuracy of 91\% on this more challenging dataset.
Using confidence thresholding, we were able to obtain a coverage of 83\% with 99\% accuracy, or
89\% coverage at 98\% accuracy. On this task, due to the larger amount of training data, we did
not see significant overfitting like we saw in SVHN so we did not use dropout. Dropout tends to
increase training time, and our largest models are already very costly to train. We also did not use maxout units. All hidden units were rectifiers
~\citep{Jarrett-ICCV2009,Glorot+al-AI-2011}.
Our best architecture for this dataset is similar to the best architecture for the public dataset,
except we use only five convolutional layers rather than eight. (We have not tried using eight convolutional
layers on this dataset; eight layers may obtain slightly better results but the version of the network with five convolutional layers
performed accurately enough to meet our business objectives) The locally connected layers have
128 units per spatial location, while the fully connected layers have 4096 units per layer.

\subsection{CAPTCHA puzzles dataset}
\label{subsec:captchas}

\begin{figure}
\begin{centering}
\includegraphics[width=0.5\textwidth]{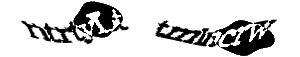}\includegraphics[width=0.5\textwidth]{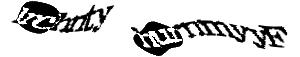}\\
\includegraphics[width=0.5\textwidth]{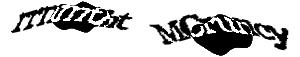}\includegraphics[width=0.5\textwidth]{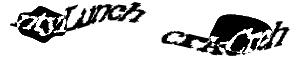}
\caption{Examples of images from the hard CAPTCHA puzzles dataset.}
\label{fig:captcha}
\end{centering}
\end{figure}

CAPTCHAs are reverse turing tests designed to use distorted text to distinguish humans and machines running automated text recognition software. reCAPTCHA is a leading CAPTCHA provider with an installed base of several hundreds of thousands of websites. To evaluate the generality of the proposed approach to recognizing arbitrary text, we created a dataset composed of the  hardest CAPTCHA puzzle examples of which are shown in Figure~\ref{fig:captcha}. 

The model we use is similar to the best one used over the SVHN dataset with the following differences: we use 9 convolutional layers in this network instead of 11, with the first layer containing normal rectifier units instead of maxouts, the convolutional layers are also slightly bigger, while the fully connected ones smaller. The output of this model is case-sensitive and it can handle up to 8 character long sequences. The input is one of the two CAPTCHA words cropped to a size of 200x40 where random sub-crops of size 195x35 are taken. The performance reported was taken directly from a test set of 100K samples and a training set in the order of millions of CAPTCHA images.

With this model, we are able to achieve a 99.8\% accuracy on transcribing the hardest reCAPTCHA puzzle. It is important to note that these results do not indicate a reduction in the anti-abuse effectiveness of reCAPTCHA as a whole. reCAPTCHA is designed to be a risk analysis engine taking a variety of different cues from the user to make the final determination of human vs bot. Today distorted text in reCAPTCHA serves increasingly as a medium to capture user engagements rather than a reverse turing in and of itself. These results do however indicate that the utility of distorted text as a reverse turing test by itself is significantly diminished.  

\subsection{Performance analysis}
\label{subsec:analysis}

\begin{figure}
\begin{centering}
\includegraphics[width=0.8\textwidth]{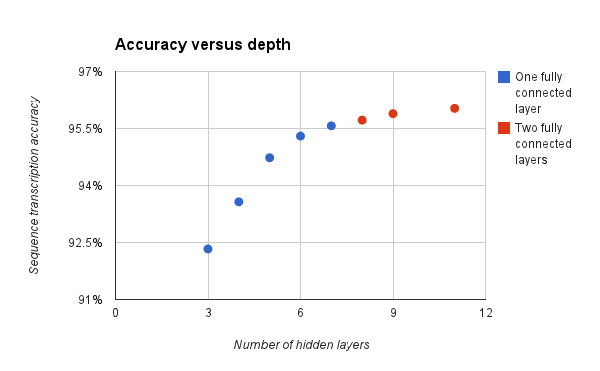}
\caption{Performance analysis experiments on the public SVHN dataset show that fairly deep
architectures are needed to obtain good performance on the sequence transcription task.}
\label{fig:depth}
\end{centering}
\end{figure}

\begin{figure}
\begin{centering}
\includegraphics[width=0.8\textwidth]{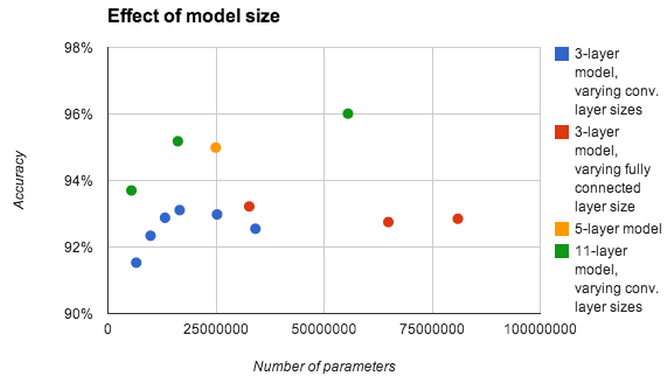}
\caption{Performance analysis experiments on the public SVHN dataset show
that increasing the number of parameters in smaller models does not allow
such models to reach the same level of performance as deep models. This is
primarily due to overfitting.}
\label{fig:model_size}
\end{centering}
\end{figure}

In this section we explore the reasons for the unprecedented success of our neural network
architecture for a complicated task involving localization and segmentation rather than just
recognition. We hypothesize that for such a complicated task, depth is crucial to achieve an
efficient representation of the task. State of the art recognition networks for images of
cropped and centered digits or objects may have between two to four
convolutional layers followed by one or two densely connected hidden layers and the classification
layers \citep{Goodfellow-et-al-ICML2013}. In this work we used several more convolutional layers.
We hypothesize that the depth was crucial to our success. This is most likely because the earlier
layers can solve the localization and segmentation tasks, and prepare a representation that has
already been segmented so that later layers can focus on just recognition.
Moreover, we hypothesize
that such deep networks have very high representational capacity, and thus need a large amount of
data to train successfully.
Prior to our successful demonstration of this system, it would have
been reasonable to expect that factors other than just depth would be necessary
to achieve good performance on these tasks. For example, it could have been possible that
a sufficiently deep network would be too difficult to optimize. In Fig.~\ref{fig:depth}, we present the results of an experiment that confirms our hypothesis that
depth is necessary for good performance on this task. For control experiments
showing that large shallow models cannot achieve the same performance, see
Fig.~\ref{fig:model_size}.

\subsection{Application to Geocoding}

The motivation for the development of this model was to decrease the cost of geocoding as well
as scale it worldwide and keep up with change in the world.
The model has now reached a high enough quality level that we can automate the extraction of
street numbers on Street View images. Also, even if the model can be considered quite large, it is still
efficient.

\begin{figure}[hb]
\begin{centering}
\makebox[\textwidth]{
  \includegraphics[width=8.0in]{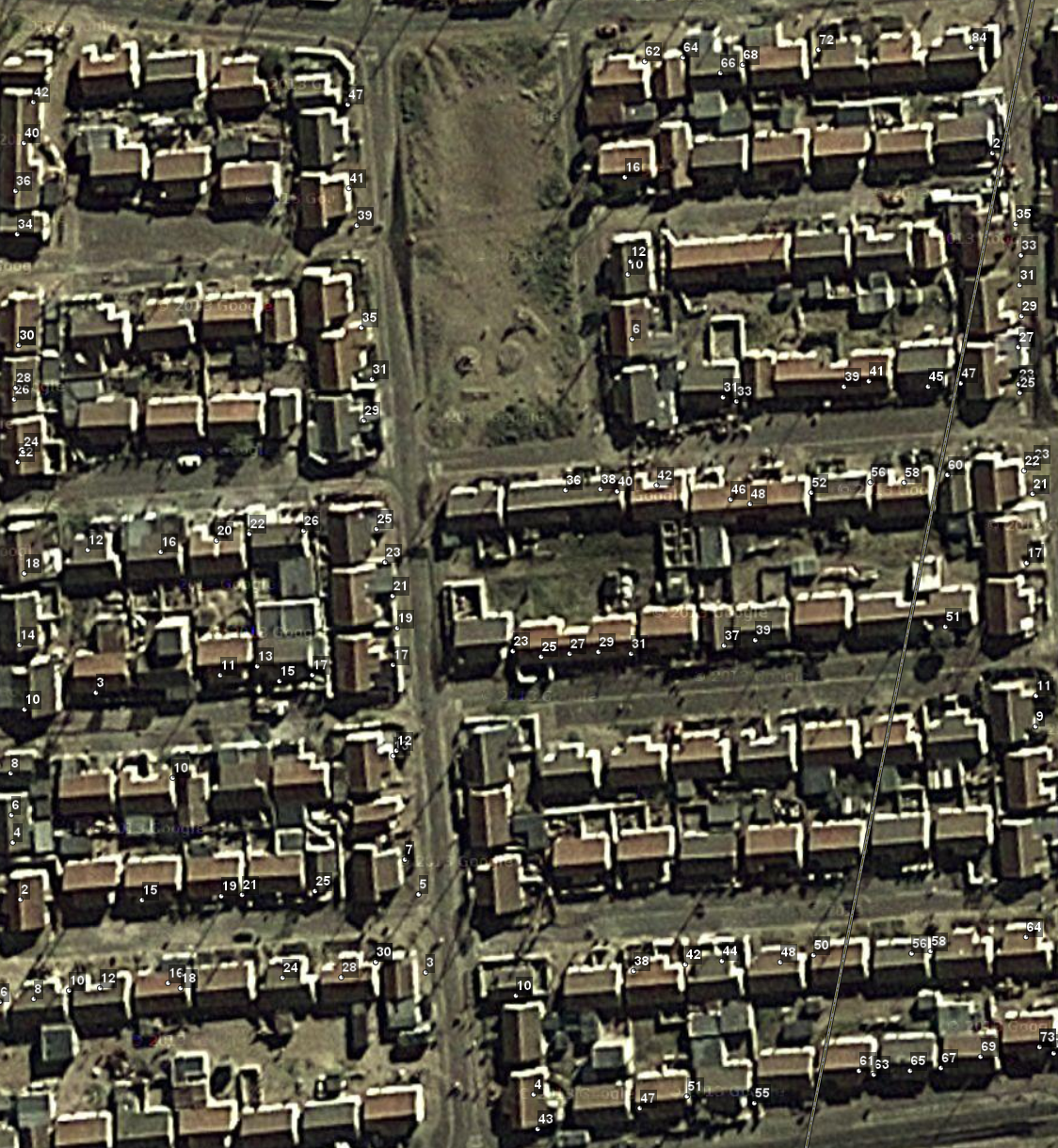}
}
\caption{Automatically extracted street numbers from Street View imagery captured in South Africa.}
\label{fig:geocoding}
\end{centering}
\end{figure}

We can for example transcribe all the views we have of street numbers in France
in less than an hour using our Google infrastructure. Most of the cost actually
comes from the detection stage that locates the street numbers in the large Street View images.
Worldwide, we automatically detected and transcribed close to 100 million physical street numbers
at operator level accuracy. Having this new dataset significantly increased the geocoding quality
of Google Maps in several countries especially the ones that did not already have other sources
of good geocoding. In Fig.~\ref{fig:geocoding}, you can see some automatically extracted street
numbers from Street View imagery captured in South Africa.

\section{Discussion}
\label{sec:discussion}

We believe with this model we have solved OCR for short sequences for many applications.
On our particular task, we believe that now the biggest gain we could easily get is to
increase the quality of the training set itself as well as increasing its size for general OCR transcription.

One caveat to our results with this architecture is that they rest heavily on the assumption that
the sequence is of bounded length, with a reasonably small maximum length $N$.
For unbounded $N$, our method is not directly applicable, and for large $N$ our
method is unlikely to scale well. Each separate digit classifier requires its own
separate weight matrix. For long sequences this could incur too high of a memory
cost. When using DistBelief, memory is not much of an issue (just use more machines)
but statistical efficiency is likely to become problematic. Another problem with long sequences is the
cost function itself. It's also possible that, due to longer sequences having more
digit probabilities multiplied together, a model of longer sequences could have trouble
with systematic underestimation of the sequence length.

One possible solution could be
to train a model that outputs one ``word'' ($N$ character sequence) at a time and then slide it over
the entire image followed by a simple decoding. Some early experiments in this direction have been
promising.

Perhaps our most interesting finding is that neural networks can learn to perform
complicated tasks such as simultaneous localization and segmentation of ordered
sequences of objects. This approach of using a single
neural network as an entire end-to-end system could be applicable to other problems,
such as general text transcription or speech recognition.

\clearpage

\subsubsection*{Acknowledgments}

We would like to thank Ilya Sutskever and Samy Bengio for helpful discussions.
We would also like to thank the entire operation team in India that did the labeling effort and without whom this research would not have been possible.
\small
\bibliography{transcription}
\bibliographystyle{natbib}

\section*{Appendix A: Example inference}

In this appendix we provide a detailed example of how to run inference in a trained network
to transcribe a house number. The purpose of this appendix is to remove any ambiguity from
the more general description in the main text.

\begin{figure}[hb]
\begin{centering}
\makebox[\textwidth]{
  \includegraphics[width=6.0in]{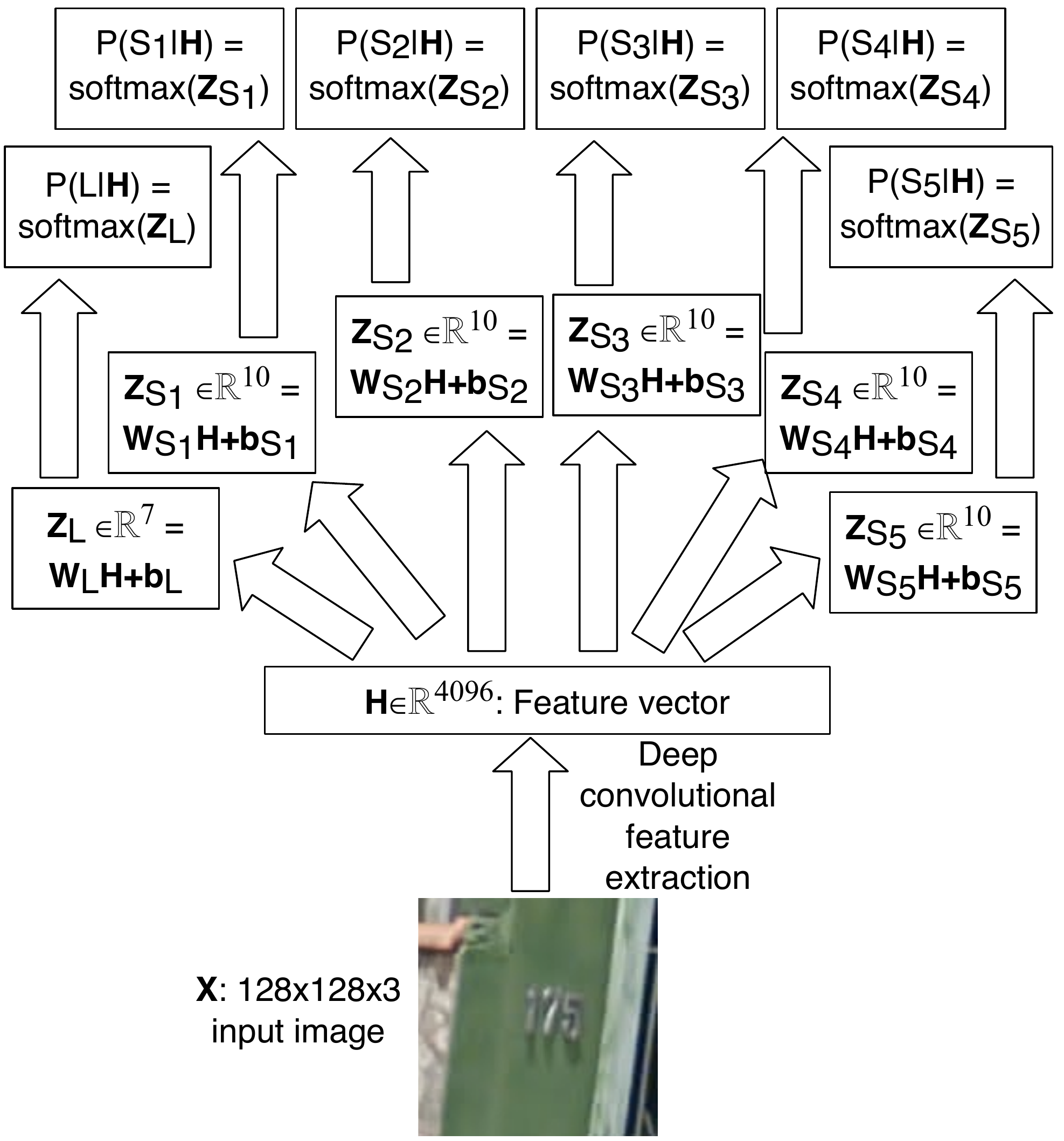}
}
\caption{Details of the computational graph we used to transcribe house numbers.
In this diagram, we show how we compute the parameters of $P( \mathbf{S} \mid \mathbf{X})$,
where $X$ is the input image and $\mathbf{S}$ is the sequence of numbers depicted by
the image. We first extract a set of features $\mathbf{H}$ from $\mathbf{X}$ using a convolutional
network with a fully connected final layer. Note that only one such feature vector is extracted
for the entire image. We do not use an HMM that models features explicitly extracted at separate
locations. Because the final layer of the convolutional feature extractor is fully connected and
has no weight sharing, we have not explicitly engineered any concept of spatial location into this
representation. The network must learn its own means of representing spatial location in $\mathbf{H}$.
Six separate softmax classifiers are then connected
to this feature vector $\mathbf{H}$, i.e., each softmax classifier forms a response by making an
affine transformation of $\mathbf{H}$ and normalizing this response with the softmax function.
One of these classifiers provides the distribution over the sequence length $P(L \mid \mathbf{H})$,
while the others provide the distribution over each of the members of the sequence,
$P(S_1 \mid \mathbf{H}), \dots, P(S_5 \mid \mathbf{H})$.
}
\label{fig:detailed_mlp}.
\end{centering}
\end{figure}

Transcription begins by computing the distribution over the sequence $\mathbf{S}$ given an
image $\mathbf{X}$. See Fig.~\ref{fig:detailed_mlp} for details of how this computation is
performed.

To commit to a single specific sequence transcription, we need to compute
$\text{argmax}_\mathbf{s} P(\mathbf{S} = \mathbf{s} \mid \mathbf{H} )$.
It is easiest to do this in log scale, to avoid multiplying together many
small numbers, since such multiplication can result in numerical underflow.
i.e., in practice we actually compute
$\text{argmax}_\mathbf{s} \log P(\mathbf{S} = \mathbf{s} \mid \mathbf{H} )$.

Note that $\log \text{softmax}( \mathbf{z})$ can be computed efficiently and
with numerical stability with the formula
$\log \text{softmax} (\mathbf{z} )_i = z_i - \sum_j \exp(z_j).$ It is best to
compute the log probabilities using this stable approach, rather than first
computing the probabilities and then taking their logarithm. The latter approach
is unstable; it can incorrectly yield $- \infty$ for small probabilities.

Suppose that we have all of our output probabilities computed, and that they
are the following (these are idealized example values, not actual values from the model):

\begin{tabular} {c|c|c|c|c|c|c|c}
& $L=0$ & $L=1$ & $L=2$ & $L=3$ & $L =4$ & $L = 5$ & $L > 5$ \\
\hline
$P(L)$ & .002 & .002 & .002 & .9 & .09 & .002 & .002 \\
\hline
$\log P(L)$ & -6.2146 & -6.2146 & -6.2146 & -0.10536 & -2.4079 & -6.2146 & -6.2146
\end{tabular}

{\tiny
\begin{tabular} {c|c|c|c|c|c|c|c|c|c|c}
& $i=0$ & $i=1$ & $i=2$ & $i=3$ & $i=4$ & $i=5$ & $i=6$ & $i=7$ & $i=8$ & $i=9$ \\
\hline
$P( S_1 = i)$ & .00125 & .9 & .00125 & .00125 & .00125 & .00125 & .00125 & .1 & .00125 & .00125 \\
\hline
$\log P( S_1 = i)$ & -6.6846 & -0.10536 & -6.6846 & -6.6846 & -6.6846 & -6.6846 & -6.6846 & -2.4079 & -6.6846 & -6.6846 \\
\hline
$P( S_2 = i)$      & .00125  & .00125   & .00125  & .00125  & .00125  & .00125  & .00125  & .9 & .00125 & .1 \\
\hline
$\log P( S_2 = i)$ & -6.6846 & -6.6846 & -6.6846 & -6.6846 & -6.6846 & -6.6846 & -6.6846 & -0.10536 & -6.6846 & -2.4079 \\
\hline
$P( S_3 = i)$     & .00125   & .00125  & .00125  & .00125  & .00125  & .9     & .1       & .00125   & .00125 & .00125 \\
\hline
$\log P( S_3 = i)$ & -6.6846 & -6.6846 & -6.6846 & -6.6846 & -6.6846 & -0.10536 & -2.4079 & -6.6846 & -6.6846 & -6.6846 \\
\hline
$P( S_4 = i)$ & .08889 & .2 & .08889 & .08889 & .08889 & .08889 & .08889 & .08889 & .08889 & .08889 \\
\hline
$\log P(S_4 = i) $ & -2.4204 & -1.6094  & -2.4204  & -2.4204  & -2.4204  & -2.4204  & -2.4204  & -2.4204  & -2.4204  & -2.4204 \\
\hline
$P( S_5 = i)$ & .1 & .1 & .1 & .1 & .1 & .1 & .1 & .1 & .1 & .1 \\
\hline
$\log P( S_5 = i)$ & -2.3026 & -2.3026 & -2.3026 & -2.3026 & -2.3026 & -2.3026 & -2.3026 & -2.3026 & -2.3026 & -2.3026
\end{tabular}
}

Refer to the example input image in Fig.~\ref{fig:detailed_mlp} to understand these probabilities. The correct length is 3.
Our distribution over $L$ accurately reflects this, though we do think there is a reasonable possibility that $L$ is 4--maybe
the edge of the door looks like a fourth digit. The correct transcription is $175$, and we do assign these digits the
highest probability, but also assign significant probability to the first digit being a $7$, the second being a $9$, or the third
being a $6$. There is no fourth digit, but if we parse the edge of the door as being a digit, there is some chance of it being a $1$.
Our distribution over the fifth digit is totally uniform since there is no fifth digit.

Our independence assumptions mean that when we compute the most likely sequence, the choice of which digit appears in each position
doesn't affect our choice of which digit appears in the other positions. We can thus pick the most likely digit in each position
separately, leaving us with this table:

\begin{tabular}{c|c|c}
$j$ & $\text{argmax}_{s_j} \log P(S_j=s_j)$ & $\text{max}_{s_j} \log P(S_j=s_j)$ \\ 
\hline
1 & 1 & -0.10536 \\
\hline
2 & 7 & -0.10536 \\
\hline
3 & 5 & -0.10536 \\
\hline
4 & 1 & -1.6094 \\
\hline
5 & 0 & -2.3026
\end{tabular}

Finally, we can complete the maximization by explicitly calculating the probability of all seven possible sequence lengths:

\begin{tabular}{c|c|c|c}
$L$ & Prediction & $\log P(S_1, \dots S_L)$ & $\log P(\mathbf{S})$ \\
\hline
0 &              & 0.      & -6.2146 \\
\hline
1 & 1            & -0.1054 & -7.2686 \\
\hline
2 & 17           & -0.2107 & -8.3226 \\
\hline
3 & 175          & -0.3161 & -0.42144 \\
\hline
4 & 1751         & -1.9255 & -4.3334 \\
\hline
5 & 17510        & -4.2281 & -10.443 \\
\hline
$>5$&17510$\dots$& -4.2281 & -10.443
\end{tabular}

Here the third column is just a cumulative sum over $\log P(S_L)$ so it can be computed in linear time.
Likewise, the fourth column is just computed by adding the third column to our existing $\log P(L)$ table.
It is not even necessary to keep this final table in memory, we can just use a for loop that generates it
one element at a time and remembers the maximal element.

The correct transcription, 175, obtains the maximal log probability of $-0.42144$, and the model outputs
this correct transcription.

\end{document}